\pdfoutput=1

\documentclass[11pt]{article}

\usepackage[preprint]{acl}

\usepackage{times}
\usepackage{latexsym}

\usepackage[T1]{fontenc}

\usepackage[utf8]{inputenc}

\usepackage{microtype}

\usepackage{inconsolata}

\usepackage{graphicx}
\usepackage{multirow}
\usepackage{booktabs} 
\usepackage{amsmath}

\title{Unveiling the Impact of Multi-Modal Interactions on User Engagement: A Comprehensive Evaluation in AI-driven Conversations}

\footnotesize{
\author{
Lichao Zhang$^{*,2}$, Jia Yu$^{*,1,2}$, Shuai Zhang$^{*,1,2}$, Long Li$^{1}$, Yangyang Zhong$^{1}$\\
\textbf{Guanbao Liang$^{1}$, Yuming Yan$^{3}$, Qing Ma$^{3}$, Fangsheng Weng$^{3}$}\\
\textbf{Fayu Pan$^{3}$, Jing Li$^{3}$, Renjun Xu$^{1}$, Zhenzhong Lan$^{\dagger,2,3}$}\\
\normalfont{$^1$Zhejiang University}, {$^2$Westlake University}, $^{3}$Westlake Xinchen Technology Co. Ltd  
}

\begin{document}
\maketitle
\begin{abstract}
Large Language Models (LLMs) have significantly advanced user-bot interactions, enabling more complex and coherent dialogues. However, the prevalent text-only modality might not fully exploit the potential for effective user engagement. This paper explores the impact of multi-modal interactions, which incorporate images and audio alongside text, on user engagement in chatbot conversations. We conduct a comprehensive analysis using a diverse set of chatbots and real-user interaction data, employing metrics such as retention rate and conversation length to evaluate user engagement. Our findings reveal a significant enhancement in user engagement with multi-modal interactions compared to text-only dialogues. Notably, the incorporation of a third modality significantly amplifies engagement beyond the benefits observed with just two modalities. These results suggest that multi-modal interactions optimize cognitive processing and facilitate richer information comprehension. This study underscores the importance of multi-modality in chatbot design, offering valuable insights for creating more engaging and immersive AI communication experiences and informing the broader AI community about the benefits of multi-modal interactions in enhancing user engagement.
\end{abstract}

\begin{figure*}[!t]
    \centering
    \includegraphics[width=.95\linewidth]{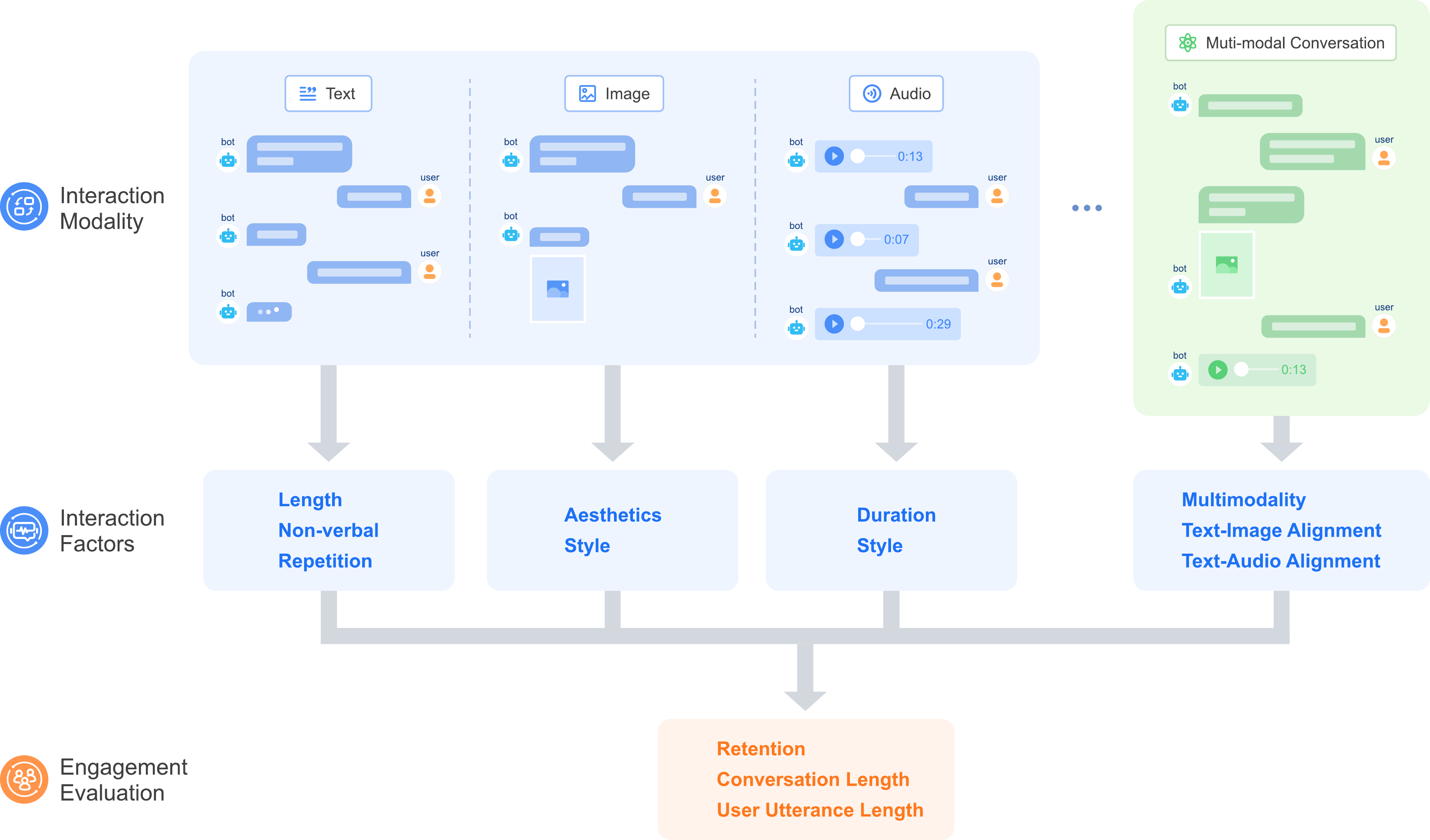}
    \caption{The overview of our comprehensive benchmark evaluating user engagement during multi-modal conversation interaction. The top row shows conversation interaction modalities, including three exemplar individual modalities (text, image and audio) conversations and an exemplar multi-modal conversation; The second row presents the conversation interaction factors derived from all four types of modalities conversations. Finally, we use three metrics to measure the user engagement with respect to the aforementioned conversation interaction factors.}
    \vspace{-3mm}
    \label{fig:enter-label}
\end{figure*}

{
\renewcommand{\thefootnote}{\fnsymbol{footnote}}
\footnotetext[1]{
 Equal contribution.
$^\dagger$Corresponding author.
}
}

\section{Introduction}
The evolution of Large Language Models (LLMs)~\cite{brown2020language,zhu2022simple} is notably enhancing the capabilities of user-bot interactions~\cite{almansor2020survey,caldarini2022literature,chaurasia2023conversational,lee2022evaluating,croes2023your}. These advanced models possess a remarkable talent for producing fluent and coherent dialogues, particularly when prompted or inspired by user inputs. Despite their proficiency, it is crucial to acknowledge that without engaging content or stimulating prompts, these bots may struggle to captivate users' attention effectively. This challenge is evident in user retention metrics, which serve as a testament to the necessity of continuous improvement in the field of conversational AI. Moreover, they also point out that the lack of sufficient user-chat interaction data limits the evaluation of user engagement.

Recently, related works~\cite{glas2015definitions,glas2015user,see2019goodconversation,ghazarian2020predictive,deriu2021survey,irvine2023rewarding,zhang2024unveiling,zheng2024judging} highlight the significant progress in leveraging LLMs to improve chatbot performance and user engagement. They emphasize the need for a holistic evaluation framework that goes beyond traditional benchmarks and considers the nuanced aspects of human-LLM interactions. 
However, these works have concentrated solely on single-modality factors to engage users.
Several studies \cite{jenkins2007analysis,kuratadevelopment,pellet2023multimodal} incorporate multi-modal features such as images and audio in interactions, marking a substantial advancement in chatbot technology. However, they still use rudimentary interface designs or employ manually crafted methods.

In our study, we address the limitation of insufficient user-bot interaction data by accumulating data from a total of 146,179 real users across various demographics including age, gender and country, and 198 bot characters with 8 categories, on our chatbot platform. This yields 747,350 dialogues, with an average conversation length of 26.09 per dialogue.

Considering that sensorial signals from multiple modalities align more closely with human attention and interest, our chatbots incorporate multi-modal mechanisms, performing in human-like manner, to engage users intuitively.
Specifically, we utilize multi-modal factors across individual modalities, such as text, image and audio, as well as their combinations, including text-image and text-audio alignments.
Subsequently, we evaluate these multi-modal factors using various objective engagement metrics, including retention, conversation length and user utterance length, to accurately measure user engagement.

We conduct comprehensive benchmark experiments and analyse across all modalities, paving the way for standardizing engagement evaluations and gaining insights on engagement effectiveness.
We have gathered consistent evidences that multi-modal factors  enhance performance significantly. The overall results indicate that conversations with multimodality featuring (scores of 0.139 in retention, 28.97 in conversation length, and 13.16 in user utterance length) are superior to text-only conversations, which scored 0.105, 15.77, and 13.01 in these metrics respectively. Moreover, our findings reveal that all non-text modalities, particularly images and audio, significantly enhance user engagement. Additionally, the integration of multiple modalities can further amplify these improvements.

Our contributions can be summarized as follows:

\begin{itemize}
\vspace{-3mm}
\item We have collected a substantial dataset comprising interactions from 146,179 real users, totaling 747,350 dialogues, with an average of 26.09 conversation length per dialogue.
\vspace{-3mm}
\item We propose to use multi-modal factors, including \textit{Length}, \textit{Non-verbal} and \textit{Repetition} in Text modality, \textit{Aesthetic} and \textit{Style} in Image modality, \textit{Duration} and \textit{Style} in Audio modality, and \textit{Multimodal Existence}, \textit{Text-Image alignment} and \textit{Text-Audio alignment} in Multimodality for comprehensively unveiling their influences on user engagement.
\vspace{-3mm}
\item We propose a comprehensive benchmark for evaluating the user engagement, including \textit{Retention}, \textit{User Utterance Length} and \textit{Conversation Length} with respect to multi-modal factors in our dataset.
\vspace{-3mm}
\item We have conducted extensive experiments on the proposed benchmark. We achieve best engagement performances, \textit{i.e.} 0.261 Average Retention, 51.53 Conversation Length and 29.55 User Utterance Length with respect to the multi-modal factor (Text-audio Alignment). We obtain substantial evidences that multi-modal factors continually improve the user engagement performance.
\end{itemize}


\section{Related Work}

\subsection{Chatbots}
LLMs have revolutionized chatbot technology, enhancing user interaction with natural, context-aware conversations~\cite{brown2020language,zhu2022simple,almansor2020survey,caldarini2022literature,chaurasia2023conversational,lee2022evaluating,croes2023your}. Further research needs to explore linguistic features and establish a standardized chatbot evaluation framework~\cite{almansor2020survey}. Current chatbots often fail to mimic human interaction due to inadequate dialogue modeling and domain-specific data~\cite{caldarini2022literature}. Ethical considerations and sector-specific applications of NLP-powered chatbots are also critical~\cite{chaurasia2023conversational}. Chatbots should enhance social communication skills and develop a `Theory of Mind' to improve human-chatbot relationships~\cite{croes2023your}. Evaluations of LLMs should incorporate interactive data instead of relying solely on standalone metrics~\cite{lee2022evaluating}. To address the lack of interactive data in previous works, our study collects extensive real-user interaction data via our chatbot platform.

\subsection{User Engagement Evaluation}
Recent research~\cite{glas2015definitions,glas2015user,see2019goodconversation,ghazarian2020predictive,deriu2021survey,irvine2023rewarding,zhang2024unveiling,zheng2024judging} has underscored the importance of controllable conversation attributes in user-chatbot engagement. This includes defining engagement behaviors~\cite{glas2015definitions}, exploring the influence of repetition, specificity, and question-asking~\cite{see2019goodconversation}, and introducing metrics like Predictive Engagement~\cite{ghazarian2020predictive}. Despite these advances, evaluations have been limited, often utilizing small participant samples~\cite{glas2015user} or focusing on single modality factors~\cite{irvine2023rewarding,zhang2024unveiling,zheng2024judging}. Comprehensive evaluation frameworks are needed~\cite{deriu2021survey}. Addressing these gaps, our study investigates multi-modal factors influencing user retention in chatbot interactions.

\subsection{Multi-modal Interactions in Chatbots}
Several studies have incorporated multi-modal interactions, including images and audio~\cite{jenkins2007analysis,kuratadevelopment,pellet2023multimodal}, but often rely on basic interfaces or hand-crafted approaches. \citet{jenkins2007analysis} use a visual component in a service-focused chatbot to encourage user engagement, but this is limited to the user interface. \citet{kuratadevelopment} develop scales to measure engagement in language learning with multimodal systems, but their questionnaire-based method increases assessment cost. ~\citet{pellet2023multimodal} present a predictive model for engagement using multimodal features, but still rely on traditional machine learning methods.

Our work advances these efforts by automatically evaluating user engagement using diverse factors from multi-modal interactions, highlighting the importance of multi-modal information in enhancing chatbot experiences.

\begin{table}[t]
\centering
\small
\begin{tabular}{@{}cccc@{}}
\toprule
\textbf{Category}                     & \textbf{User} & \textbf{Bot}       \\ \midrule
\#Speakers                            & $146,179$     & $198$              \\
\#Avg. text utterance per speaker     & $26.33$       & $26.61$            \\
\#Avg. image utterance per speaker    & -             & $3.02$             \\
\#Avg. audio utterance per speaker    & -             & $4.30$             \\
\bottomrule 
\end{tabular}
\caption{Statistics of all dialogs. Total number of bots, corresponding user count, and average rounds of text, audio, and image use per conversation.}
\vspace{-5mm}
\label{tab:overall_stat}
\end{table}
\section{Data Sourcing and Statistical Analysis}
\subsection{Platform Setting and Dataset Collection}
Our AI-chat platform provides a unique and personalized role-playing experience, fostering deep emotional connections. It supports interactions with numerous predefined characters and also offers character customization to suit user preferences, including name, gender, personality, and appearance. This personalized interaction with anthropomorphized characters cultivates intimate emotional bonds.
Our primary goal is to enable flexible chat interactions without direct incentives, except for daily login rewards.
Users can freely choose their preferred communication method, such as image requests or voice calls, while we boost engagement by sharing context-appropriate images based on conversation depth.

\begin{figure}[!th]
    \centering
    \includegraphics[width=\linewidth]{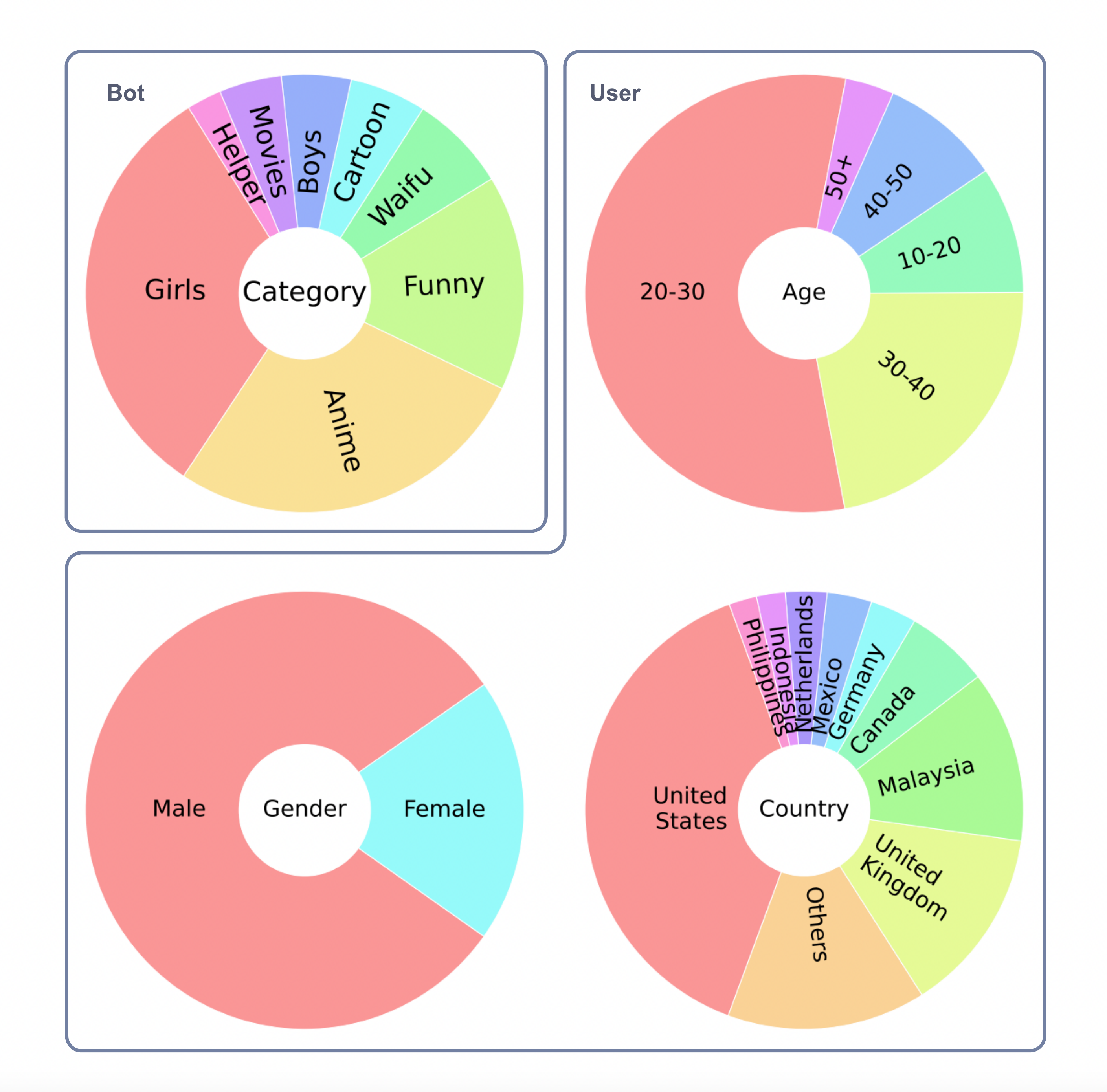}
    \caption{Proportions of different types of characters and users. The top left illustrates the proportions of different character types, the top right shows the age distribution of users, the bottom left indicates the gender distribution among users, and the bottom right displays the geographic distribution of users.}
    \label{fig:demographic}
    \vspace{-4mm}
\end{figure}

We have amassed an extensive collection of real role-playing dialogue data from our platform, which boasts a diverse array of role-playing characters and a substantial user base. This dataset encompasses interactions between actual users and characters portrayed by role-playing models. These characters can send messages to users in various formats, including text, images, and audio. Our objective is to examine whether the underlying factors of these messages can influence user engagement. This is achieved by comparing user behavior under different conditions of these factors, across various characters, as evidenced in the real conversation data. To ensure the statistical validity of our comparisons, we exclude charactersharacters involved in fewer than 200 dialogues. Following this filtration process, our dataset comprises a total of 146,179 real users and 198 characters (Table \ref{tab:overall_stat}).

\subsection{User-bot Speaker Statistic}
Among the total of 198 bot characters, we observe a diverse assortment, categorized into 8 types. The three most prevalent types are Girls (27.2\%), Anime (15.9\%), and Funny (7.2\%), as shown in the top left section of Figure \ref{fig:demographic}. Regarding our user base of 146,179 real users, we see a broad spectrum across different genders, with approximately a 4:1 ratio of males to females, depicted in the bottom left section of Figure \ref{fig:demographic}. The top three age groups are 20-30 years (56.0\%), 30-40 years (22.1\%), and 10-20 years (9.4\%), illustrated in the top right section of Figure \ref{fig:demographic}. Users come from over 150 countries, with the top three being the United States (38.8\%), the United Kingdom (13.7\%), and Malaysia (12.7\%), represented in the bottom right section of Figure \ref{fig:demographic}.

\subsection{Dialog Statistic}
As illustrated in Table \ref{tab:dialog_type}, our dataset comprises a total of 747,350 dialogs. This includes 354,465 text-only dialogs, 250,569 dialogs using both text and images, 48,321 dialogs using text and audio, and 93,995 dialogs employing text, images, and audio. We observe that the length of the conversation tends to increase with the use of more modalities in the dialog.


\begin{table}[t]
\scalebox{1.0}{
\begin{tabular}{@{}cccc@{}}
\toprule
\textbf{Dialog type}            & \textbf{\#Dialog}    & \textbf{\#Avg. CL}   \\ \midrule
Text                            & $354,465$            & $14.70$              \\
Text + Image                    & $250,569$            & $24.15$              \\
Text + Audio                    & $48,321$             & $44.04$              \\
Text + Image + Audio            & $93,995$             & $46.54$              \\\hline
Total                           & $747,350$            & $26.09$              \\
\bottomrule 
\end{tabular}}
\caption{Statistics of various dialog types, categorized by different modalities used in the dialog, alongside the average number of rounds per dialogue for each type.}
\vspace{-5mm}
\label{tab:dialog_type}
\end{table}

\section{Conversation Interaction Factors}
To objectively understand the mechanisms driving user retention, we engage in a quantitative assessment of user interaction preferences. This is achieved by measuring multi-modal factors that are characterized as atomic, orthogonal, and explainable. These factors are derived from sampled dialog data corresponding to each model, allowing us to calculate and analyze the specific elements that contribute to maintaining user engagement.

In this section, we systematically categorize the factors influencing user engagement according to their modalities. We address factors pertinent to `Text', which includes written content and its impact on user retention. Following this, we explore `Image' factors, detailing how visual elements contribute to user engagement. Subsequently, we consider `Audio' factors, examining the role of sound in maintaining user interest. Finally, we delve into `Multi-modal Interactions', which integrate text, image, and audio elements, to understand their combined effect on user retention.
\subsection{Text}
We use three factors potentially influence user engagement in the text modality, including length, non-verbal description and repetition.

\noindent\textbf{Length}. The length of a chatbot's response serves as a fundamental attribute that conveys information and has the potential to influence user engagement. To calculate the average number of words per utterance, we use the equation $\frac{1}{T} \sum_{t=1}^{T} l_t \nonumber$ where $l_t$ represents the word count in the $t$-th turn's response by the chatbot.




\noindent\textbf{Repetition}. Semantic repetition in user interactions can serve as an indicator of their level of interest and engagement, particularly in the context of habitual behaviors. We assess the frequency of semantically repetitive consecutive utterances within a conversation and calculate the average frequency using the equation, $\frac{1}{T-1} \sum_{t=1}^{T-1} \text{I}_{\text{repetition}}(u_t, u_{t+1})$.
The function \(\text{I}_{\text{repetition}}(u_t, u_{t+1})\) indicates whether two successive model utterances, \(u_t\) and \(u_{t+1}\), are semantically repetitive. This is determined by computing the cosine similarity of their representations, with a similarity score above 0.95 signifying semantic repetition.

\noindent\textbf{Non-verbal}. Incorporating non-verbal cues in a chatbot's responses can enhance the dialogue by depicting environmental changes and expressing characters' emotional conditions, which could potentially boost user engagement. To quantify the occurrence of utterances with non-verbal descriptors, we use the formula, 
$\frac{1}{T} \sum_{t=1}^{T} \text{I}_{\text{non-verbal}}(u_t)$.
In this equation, $\text{I}_{\text{non-verbal}}(u_t)$ is a function that checks if the utterance $u_t$ from the model contains non-verbal elements.

\subsection{Image}
We employ two factors for assessing user engagement in the image modality, i.e. the aesthetic score of and the style of the image.

\noindent\textbf{Aesthetic.} We introduce the LAION-Aesthetics-Predictor V2~\cite{LAION-Aesthetics-Predictor-V2} as an effective evaluation tool to measure images' aesthetic quality. It utilizes the CLIP ViT/14 model as a linear foundation model trained on large-scale image datasets~\cite{AVA),pressmancrowson2022}. It employs a multi-layer perceptron (MLP) model to predict the aesthetic quality of images. When an image is inputted, the model generates a score from 1 to 10, with higher scores representing greater aesthetic appeal.

\noindent\textbf{Style.} In pursuit of investigating the influence of image styles on user engagement, our conversational AI system is programmed to generate images in three distinctive styles: \textit{2.5D}, \textit{realistic}, and \textit{animated}. 
The engagement triggered by each style will be carefully examined to understand their individual effects.

\subsection{Audio}
We consider two factors that potentially affect user engagement in the dialogue: the duration and the style of the audio.

\noindent\textbf{Duration.} Audio component in the chatbot's responses to users is a significant modality detail, as many users prefer audio chats as it aligns with their preferences. 
Thus, we utilize the audio duration as a primary metric for statistically assessing user engagement with the chatbot in the audio modality.
The definition of \textit{Duration} is presented as $\frac{1}{T}\sum_{t=1}^{T} d_t$, where $d_t$ represents duration of the audio utterance in the $t$-th turn of a conversation by the chatbot.

\noindent\textbf{Style.} In a multi-modal context, the audio style in the user-bot dialogue is crucial. Each chatbot persona is assigned an audio style, classified into two distinct categories: \textit{female} and \textit{male}. 
In this scenario, the audio style of the spoken content is regarded as a key factor in evaluating user engagement within the audio modality.
\subsection{Multi-modal Interactions}
We include multi-modal factors for assessing user engagement: multimodality and alignment among different multimodal elements, including text-image and text-audio alignments in dialogues.

\noindent\textbf{Multimodality}. Whether dialogues enriched with multimodal element. Dialogues that incorporate additional modalities, such as images and audio, beyond mere text, may augment the interestingness of the interaction and potentially enhance user engagement.

\noindent\textbf{Text-image Alignment}.
During the process of image modality dialogue interaction between users and robots, it is insufficient to merely analyze the features of image and text modalities independently. If the image response provided by the robot does not align with the user's actual intent, it can significantly impact user engagement. Therefore, we have extracted dialogue data containing images from a vast number of real conversations. We select the five rounds of historical dialogue prior to the robot's image response as the representation of the user's intent. This, along with the image generated by the robot, is then inputted into the Qwen-vl~\cite{bai2023qwen}. We have designed a specific prompt to score the alignment between the image and the user's intent. The scoring range is from 1 to 10, with a higher score indicating a higher degree of alignment.

\noindent\textbf{Text-audio Alignment}.
The semantic alignment between the intent of text dialogues preceding the trigger of voice dialogues and the content of voice dialogues following the trigger might be a crucial factor affecting user engagement. We extracted the content of five rounds of text dialogues prior to the trigger of voice dialogues, along with the content of the voice dialogues. We constructed appropriate prompts and utilized Qwen1.5-32B-Chat~\cite{qwen32b} to provide a score for the degree of alignment. The scoring range is from 0 to 1, with a higher score indicating a higher degree of relevance. 

\section{User Engagement Evaluation}
To comprehensively analyze user engagement in interactions with bots, we propose to utilize three primary metrics: retention, conversation length (CL), and user utterance length (UUL).
These metrics provide a holistic perspective of user interactions, capturing elements from engagement duration to conversation complexity and continuity.
\subsection{Retention}
The primary metric that businesses often aim to improve is user retention on the chatbot platform. Day X retention represents the proportion of users who revisit the chatbot on the Xth day after their first interaction, a process that can be expensive due to the necessity of maintaining user engagement for at least X days.
Instead, we use the next-day retention rate as our method of measurement on our platform. High retention rate typically signals strong user stickiness.
The formula of Retention can be written as $\frac{1}{N} \sum_{n=1}^{N}r_n \nonumber$ where $r_n$ denotes whether the user persists into the subsequent phase (with $r_n = 1$ if user persists,  otherwise 0), across $N$ conversations.

\subsection{Conversation Length}
We use CL, which counts the interactions between user and bot, to assess user engagement. This metric provides a holistic assessment from both user and bot perspectives. It suggests that a longer conversation length indicates active user engagement and the bot's responsive adaptability.

For precise measurement, the interaction turns $T_n$ between the user and bot in $n$-th conversation are logged, excluding any monologues. Then, the average conversation length is defined as $\frac{1}{N} \sum_{n=1}^N T_n$, across $N$ conversations.

\subsection{User Utterance Length}
We introduce UUL as our evaluative criteria for analyzing user engagement. For calculating UUL, we sum the lengths of all user utterances together in a dialog and then divide by the number of user utterances corresponding, formulated as below:
\begin{equation}
UUL=\frac{1}{\sum_{n=1}^N T_n} \sum_{n=1}^{N}\sum_{t=1}^{T_n} l_{t}^n\nonumber
\end{equation}

Here, $l_{t}^n$ represents the length of the user utterance in the $t$-th turn of the $n$-th conversation.

\section{Experimental Results}

\begin{table*}[t]
\centering
\small
\begin{tabular}{|c|c|c|ccc|}
\hline
\textbf{Modalities}&\textbf{Factors} & \textbf{Condition (value)}                     & \textbf{\#Avg. Retention$\uparrow$}   & \textbf{\#Avg. CL$\uparrow$}  & \textbf{\#Avg. UUL$\uparrow$}  \\ \hline\hline
\multirow{6}{*}{Text}&\multirow{2}{*}{Length}& Lower (43.4)& 0.076                  & 11.39                 & 9.91                  \\
&&Higher (91.5)& \textbf{0.134*}        & \textbf{20.14*}       & \textbf{14.75*}       \\ \cline{2-6}
&\multirow{2}{*}{Non-verbal}&Lower (0.52)& \textbf{0.110*}        & \textbf{16.82*}       & 12.73                 \\
&&Higher (0.96)& 0.100                  & 14.72                 & \textbf{13.12}        \\\cline{2-6}
&\multirow{2}{*}{Repetition}& Lower (0.00)& 0.096                  & 13.69                 & 12.63                 \\
&& Higher (0.02)& \textbf{0.114*}        & \textbf{17.85*}       & \textbf{13.27*}       \\\hline
\multirow{5}{*}{Image}&\multirow{2}{*}{Aesthetics} & Lower (6.20)& \textbf{0.127}         & \textbf{24.79}        & 12.66                 \\
&&Higher (6.47)& 0.126                  & 24.18                 & \textbf{12.79}       \\ \cline{2-6}
&\multirow{3}{*}{Style}&2.5D                            & \textbf{0.134*}        & \textbf{26.50*}       & \textbf{13.47}        \\
&&Photorealistic                                                 & 0.126†                  & 23.77†                 & 12.58†                 \\ 
&&Anime                                                & 0.121                  & 23.25                 & 12.54                 \\ \hline
\multirow{4}{*}{Audio}&\multirow{2}{*}{Duration}&Lower (5.92)& 0.162                  & 38.04        & 12.35                 \\
&&Higher (18.10)& \textbf{0.201*}        & \textbf{50.52*}       & \textbf{14.26*}       \\ \cline{2-6}
&\multirow{2}{*}{Style}&Female                         & 0.182                  & 44.38                 & 13.46                 \\
&&Male                                                 & \textbf{0.190}         & \textbf{45.05}        & \textbf{13.76}        \\ \hline
\multirow{6}{*}{Multi-modal}&\multirow{2}{*}{Multimodality}& No                 & 0.105                  & 15.77                 & 13.01                 \\
&& Yes                                               & \textbf{0.139*}        & \textbf{28.97*}       & \textbf{13.16}        \\ \cline{2-6}
&\multirow{2}{*}{Text-image Alignment}&Lower (7)& {0.084}                     & {\textbf{27.54}}                    & {{16.36}}                    \\
&&Higher (8)& {\textbf{0.152}}                      & {24.58}                   & \textbf{19.07}                     \\ \cline{2-6}
&\multirow{2}{*}{Text-audio Alignment}&Lower (2-5)& {0.159}                  & {42.82}                & \textbf{29.55}                 \\
&&Higher (6-9)& {\textbf{0.261}}  & {\textbf{51.53}}        & {{16.46}}        \\ \hline
\end{tabular}
\caption{Comparison of the Influence of Different Modality Factors on User Engagement: User Retention, Conversation Length, and User Utterance Length. Each factor's conditions are compared (values, mean values, or score ranges are provided in parentheses). The average value for each user engagement measure is calculated, with the highest value highlighted in bold. An asterisk (*) indicates a value significantly larger than those under other conditions. A dagger (†) indicates no significant relation with other conditions.}
\vspace{-2mm}
\label{tab:factor_influence}
\end{table*}

\begin{figure}
    \centering
    \includegraphics[width=.95\linewidth]{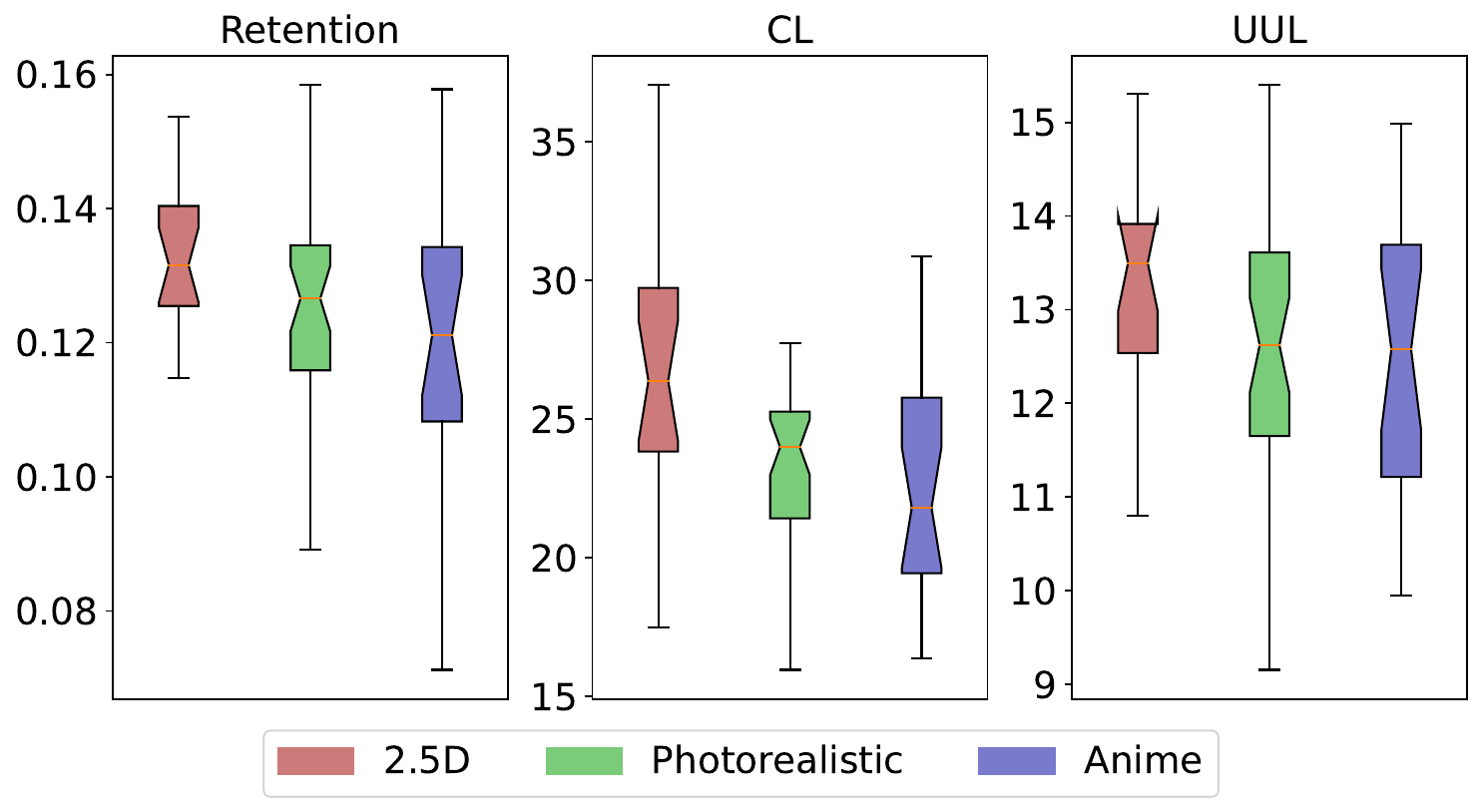}
    \caption{Comparison of the influence of different image styles on three user engagement measures. Each subfigure compares the distribution of each measure across all image styles.}
    \vspace{-2mm}
\end{figure}

\begin{figure}
    \centering
    \includegraphics[width=.95\linewidth]{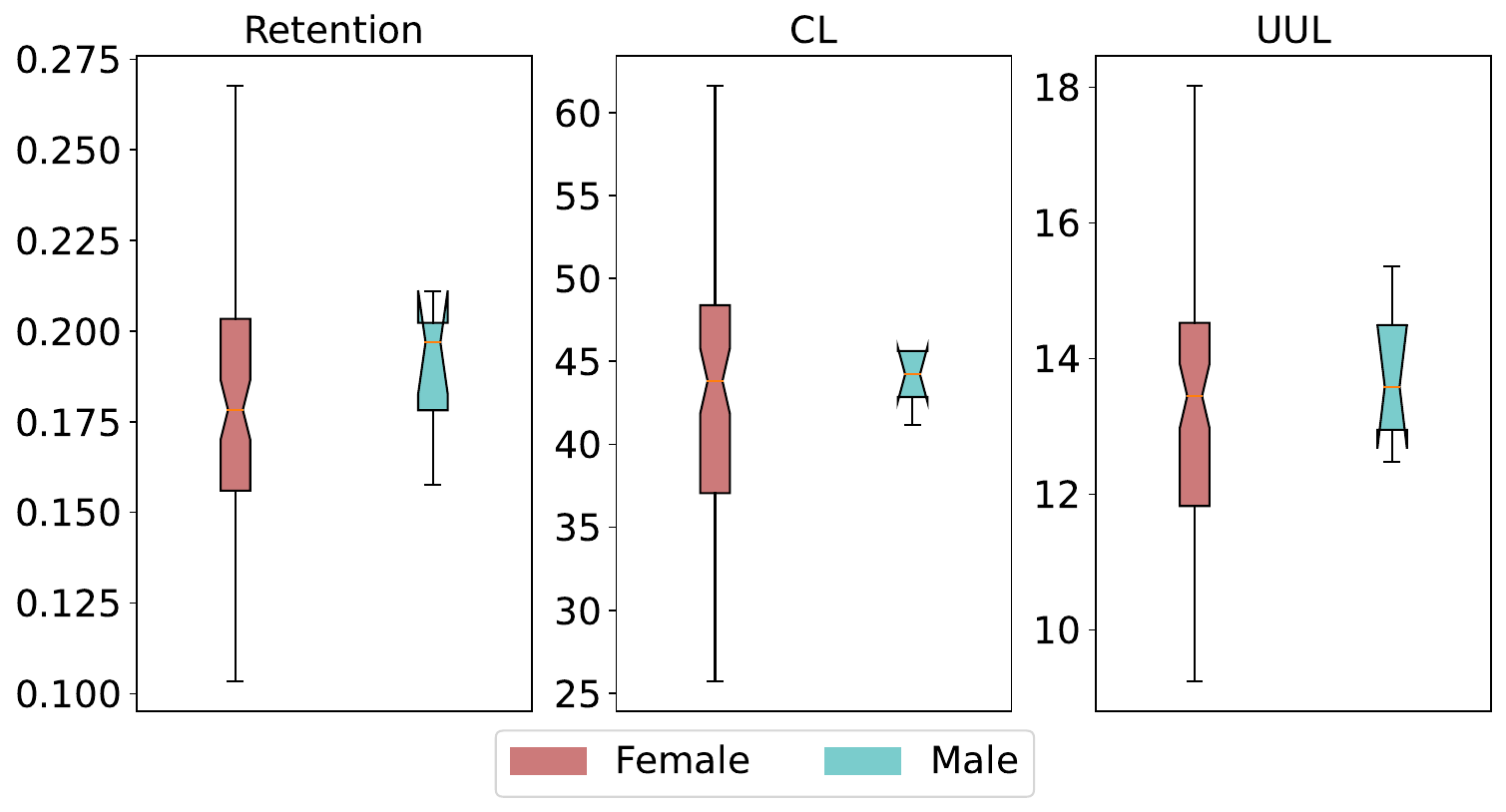}
    \caption{Comparison of the influence of different audio styles on three user engagement measures. Each subfigure compares the distribution of each measure across all audio styles.}
    \vspace{-2mm}
\end{figure}

\subsection{Setup}
We employ `Retention', `CL' and `UUL' as the main metric for evaluating user engagement. For single-modal evaluating factors, we use length, non-verbal, repetition in the text modality, aesthetics and style in the image modality, duration and style in the audio modality. For multi-modal assessment, we use multimodality, text-image alignment, and text-audio alignment.
We divide the each factor into `Lower' condition and `Higher' condition, by calculating the mean values correspondingly.

\subsection{Text Results}
\begin{figure}[!b]
    \centering
    \includegraphics[width=\linewidth]{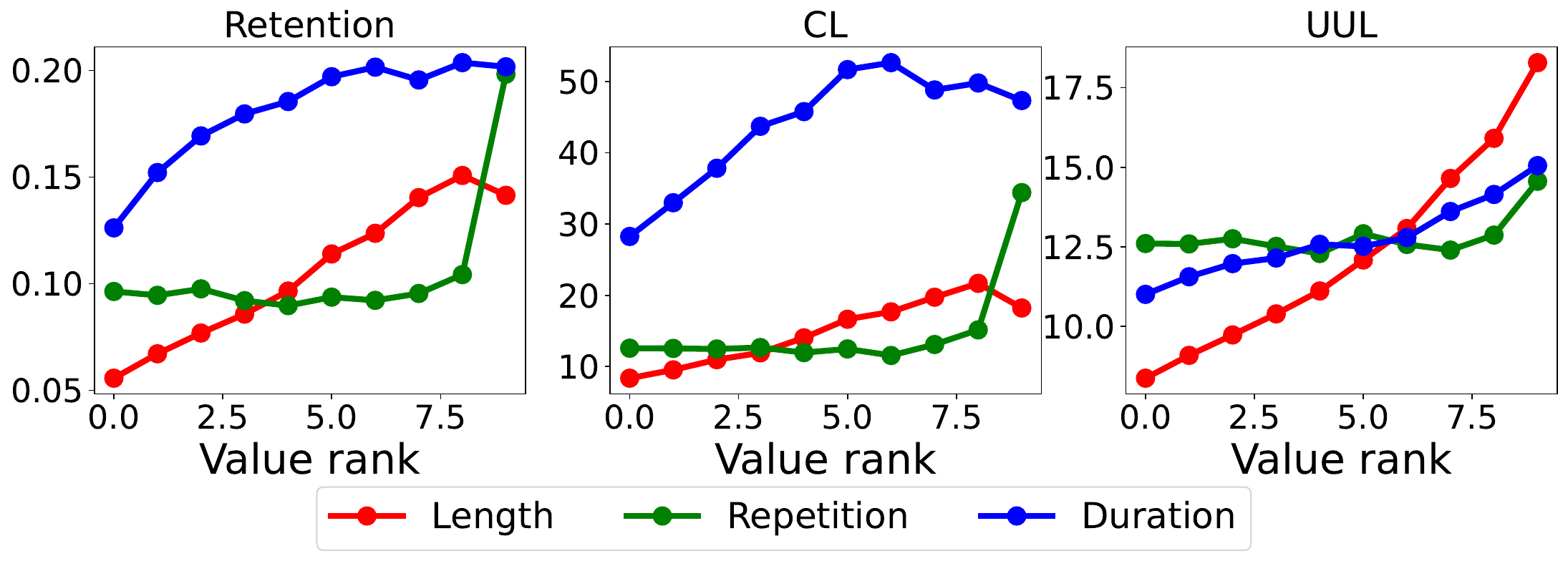}
    \caption{Single-modal correlation trends. The x-axis represents quantified condition values, while the y-axis shows Retention, CL, and UUL results. A rise in condition values corresponds to improved user engagement, highlighting a clear positive correlation.}
    \label{text_audio_combine_bins}
    \vspace{-4mm}
\end{figure}

\noindent\textbf{Length}. When comparing user behavior between longer (with a mean value of 91.5) and shorter (with a mean value of 43.4) utterances from the chatbot, we observe that longer utterances can significantly increase the retention rate, conversation length, and length of the user's utterance, thereby enhancing user engagement (Table \ref{tab:factor_influence}). Also, the higher results condition on length are superior to lower ones as in top-left of Figure~\ref{fig:factor_influence}.
Moreover, we find strict positive correlations between the text length and all metrics, i.e. retention, CL and UUL, as in Figure~\ref{text_audio_combine_bins}.

\noindent\textbf{Non-verbal}. In Table \ref{tab:factor_influence}, it demonstrates that less frequently used non-verbal descriptions do increase user retention and conversation length. 
Also, we observe that the higher and lower results condition on non-verbal are general consistency as in coordinate of (row 2, column 1) in Figure~\ref{fig:factor_influence}.

\noindent\textbf{Repetition}. Interestingly, our results reveal that, compared to non-repetitive adjacent responses from the chatbot, slight repetition can improve results of retention, CL and UUL, thereby enhancing user engagement, as in Table \ref{tab:factor_influence}. Similar results are observed in bottom left of Figure~\ref{fig:factor_influence}.
Interestingly, we observe correlation relationship between the repetition with the three metrics, i.e. retention, CL and UUL, shown in Figure~\ref{text_audio_combine_bins},

\subsection{Image Results}
\begin{figure*}[!t]
    \centering
    \includegraphics[width=.9\linewidth]{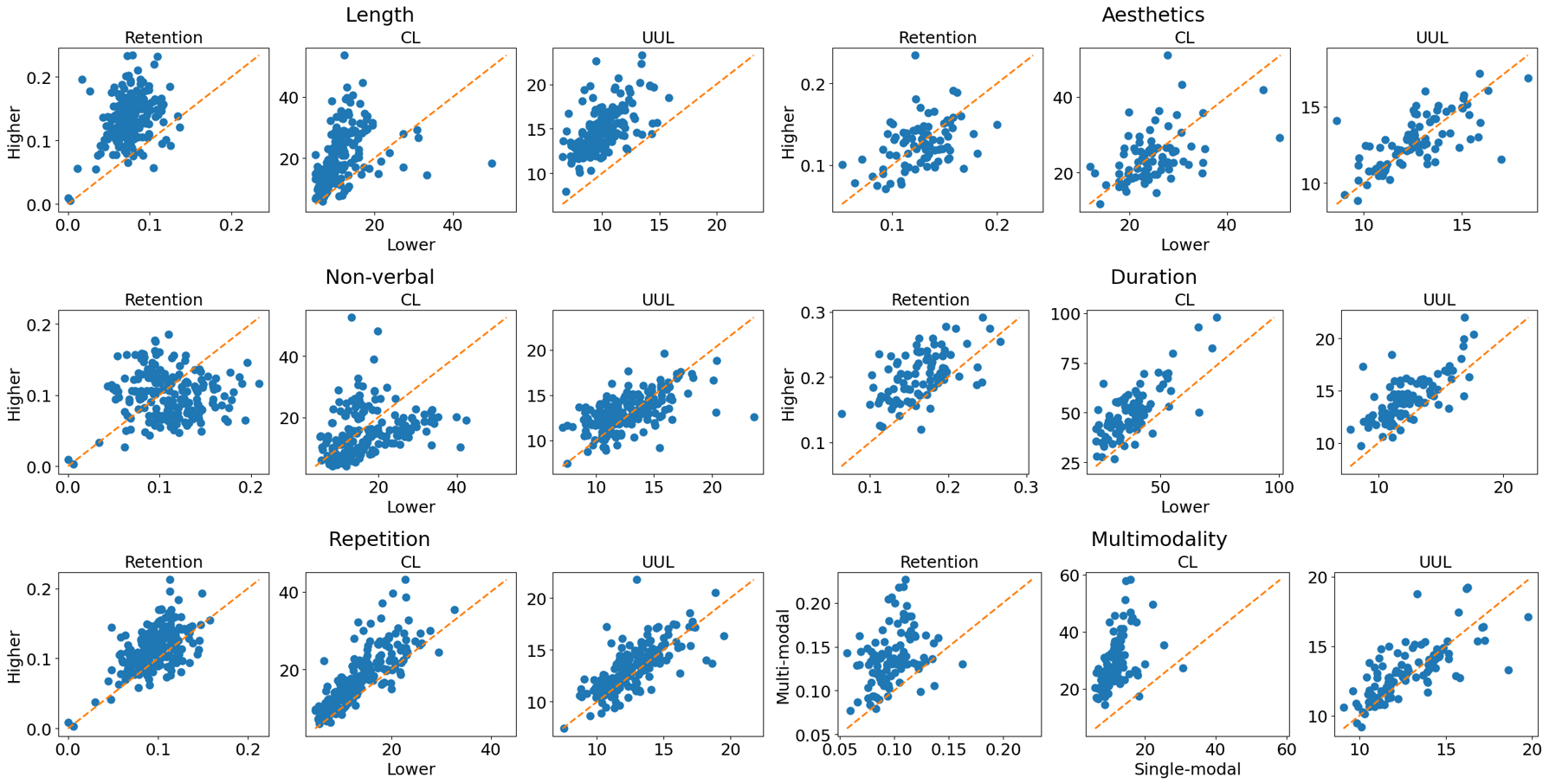}
    \caption{Detailed comparison of three user engagement measures between lower and higher values of six factors, corresponding to six groups of figures. Each group contains three measure comparisons. For each comparison of user engagement, each point represents one role-playing chatbot. The x-axis represents the mean value of this engagement in dialogues facilitated by this chatbot under conditions of lower factor values, while the y-axis represents those under conditions of higher factor values. Dashed lines are included to denote equal values under both sets of conditions.}
    \vspace{-4mm}
    \label{fig:factor_influence}
\end{figure*}

\vspace{-1.5mm}
\noindent\textbf{Aesthetics}. Table \ref{tab:factor_influence} shows minor differences between lower and higher aesthetics. The results in top right of Figure~\ref{fig:factor_influence} also demonstrate their approximate equivalence.

\noindent\textbf{Style}. In Table~\ref{tab:factor_influence}, our results demonstrate that the style of the image used in the dialogue can significantly influence user behavior. We found that 2.5D style images can significantly enhance interaction depth and duration compared with anime images, while on par with photorealistic style images.

\vspace{-2mm}
\subsection{Audio Results}
\vspace{-1.5mm}
\noindent\textbf{Duration}. Our study reveals that longer audio duration in the dialogue continuously improve results in terms of retention, CL, and UUL,  compared to shorter duration, as in Table \ref{tab:factor_influence}.
It can be seen that the higher results condition on duration are superior to lower ones as in coordinate of (row 2, column 2) in Figure~\ref{fig:factor_influence}.
Similar with length and repetition in Figure~\ref{text_audio_combine_bins}, we observe distinct correlation relationship between the repetition with the three metrics, i.e. retention, CL and UUL.

\noindent\textbf{Style}. In Table \ref{tab:factor_influence}, we find that female-voiced audio in the dialogue and male-voiced audio keep approximate equivalent results with respect to retention, CL and UUL. 
Our dataset reveals no significant impact of these two kinds of audio style on user engagement, but this doesn't rule out the potential influence of other audio styles.

\subsection{Multi-modal Results}
\vspace{-1.5mm}
\noindent\textbf{Multimodality}. Our findings indicate that compared to text-only dialogues, incorporating modalities in dialogues can enhance user engagement by improving both the interaction depth and duration, as evidenced in Table~\ref{tab:factor_influence}. 
It can be seen that the higher results condition on multimodality are superior to lower ones in terms of retention and CL. While it keeps approximate equivalence in terms of UUL in bottom right of Figure~\ref{fig:factor_influence}.

\begin{figure}[h]
    \centering
    \includegraphics[width=\linewidth]{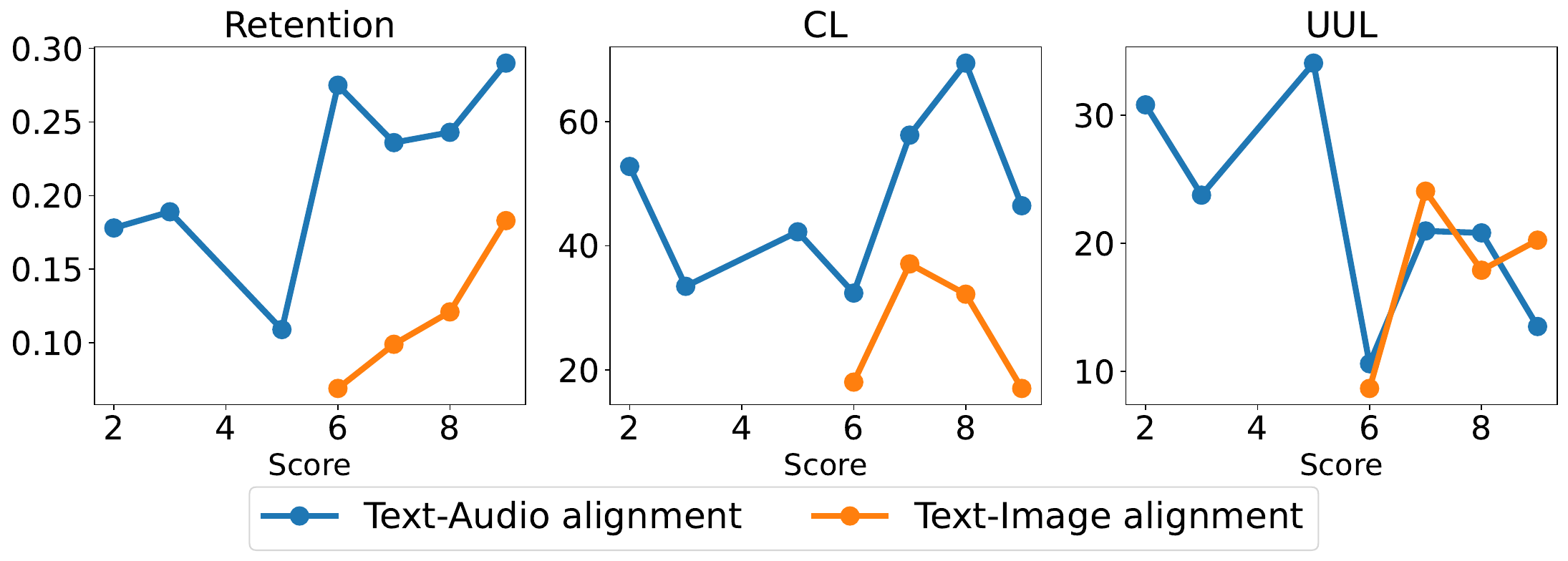}
    \caption{Correlation trends for text-audio (Blue) and text-image (Orange) alignment scores in relation to Retention, CL, and UUL respectively.}
    \label{alignment_score}
    \vspace{-4mm}
\end{figure}

\noindent\textbf{Text-image Alignment}.
In the bottom of Table~\ref{tab:factor_influence}, we compare text-image alignment conditioned on lower values (7) and higher values (8). It showcases the higher alignment score, the better performances in most cases of Retention, CL and UUL.
Importantly, we find it generally obeys the positive correlation in cases of Retention and UUL, as in Figure~\ref{alignment_score}, suggesting that the effectiveness of our proposed text-image alignment metrics.

\noindent\textbf{Text-audio Alignment}.
Similarly, we analyze the text-audio alignment in the bottom of Table~\ref{tab:factor_influence}.
We obtain results of 0.261 Retention and 51.53 CL for higher condition values, compared with 0.159 Retention and 42.82 CL for lower condition values.
It evidences that a higher degree of text-audio alignment can positively impact user engagement. 
Moreover, as shown in Figure~\ref{alignment_score}, we observe clear positive correlations for both retention and CL, strongly validating the effectiveness of our text-audio alignment measure.

\vspace{-1mm}
\section{Conclusion}
In conclusion, we tackle the issue of limited user-chat interaction data by amassing a dataset from 146,179 real users across diverse demographics on our chatbot platform, leading to 747,350 dialogues with an average of 26.09 conversation length each.
We employ multi-modal elements like text, image, and audio, including their combinations, to foster a more intuitive and human-like user engagement.

We then evaluate these multi-modal factors using objective engagement metrics such as retention, conversation length, and user utterance length. Through comprehensive benchmark experiments across all modalities, we gain insights into engagement effectiveness and standardize evaluations.
Our results consistently show that conversation interaction factors, including image, audio and multi-modal interactions, significantly boost user engagement performance in chatbot conversations.

\section{Limitations}
Despite our research making significant strides, it does come with certain limitations: 
\begin{itemize}
\item Our method of identifying factors that influence user retention is dependent on statistical correlations, which do not necessarily imply causality. Future research should focus on establishing causal links between these factors and user interaction preferences. 
\item Moreover, our methods are somewhat limited by the tools we utilized for factor quantification and calculation, including representation models or APIs. As these tools advance, they hold the potential to refine and improve our findings.
\item For the analysis of multi-modal interaction data, we should adopt more objective and quantifiable analysis methods. For instance, incorporating human evaluation can make the analysis more precise and comprehensive. In future research, we plan to include human evaluation to make our benchmark tests more scientific and effective.
\end{itemize}
\section{Ethics Statement}
The data utilized in this study originates from actual user interactions with the role-playing model. Under our terms of service, these users have given their consent for their data to be employed for scientific research. Crucially, the data we have used is devoid of any personal details, thereby safeguarding the privacy and confidentiality of our users.

\bibliography{custom}

\begin{thebibliography}{23}
\providecommand{\natexlab}[1]{#1}

\bibitem[{Almansor and Hussain(2020)}]{almansor2020survey}
Ebtesam~H Almansor and Farookh~Khadeer Hussain. 2020.
\newblock Survey on intelligent chatbots: State-of-the-art and future research directions.
\newblock In \emph{Complex, Intelligent, and Software Intensive Systems: Proceedings of the 13th International Conference on Complex, Intelligent, and Software Intensive Systems (CISIS-2019)}, pages 534--543. Springer.

\bibitem[{Bai et~al.(2023{\natexlab{a}})Bai, Bai, Yang et~al.}]{bai2023qwen}
J~Bai, S~Bai, S~Yang, et~al. 2023{\natexlab{a}}.
\newblock \href {https://openreview.net/pdf?id=qrGjFJVl3m} {Qwen-vl: A versatile vision-language model for understanding, localization, text reading, and beyond}.

\bibitem[{Bai et~al.(2023{\natexlab{b}})Bai, Bai, Chu, Cui, Dang, Deng, Fan, Ge, Han, Huang, Hui, Ji, Li, Lin, Lin, Liu, Liu, Lu, Lu, Ma, Men, Ren, Ren, Tan, Tan, Tu, Wang, Wang, Wang, Wu, Xu, Xu, Yang, Yang, Yang, Yang, Yao, Yu, Yuan, Yuan, Zhang, Zhang, Zhang, Zhang, Zhou, Zhou, Zhou, and Zhu}]{qwen32b}
Jinze Bai, Shuai Bai, Yunfei Chu, Zeyu Cui, Kai Dang, Xiaodong Deng, Yang Fan, Wenbin Ge, Yu~Han, Fei Huang, Binyuan Hui, Luo Ji, Mei Li, Junyang Lin, Runji Lin, Dayiheng Liu, Gao Liu, Chengqiang Lu, Keming Lu, Jianxin Ma, Rui Men, Xingzhang Ren, Xuancheng Ren, Chuanqi Tan, Sinan Tan, Jianhong Tu, Peng Wang, Shijie Wang, Wei Wang, Shengguang Wu, Benfeng Xu, Jin Xu, An~Yang, Hao Yang, Jian Yang, Shusheng Yang, Yang Yao, Bowen Yu, Hongyi Yuan, Zheng Yuan, Jianwei Zhang, Xingxuan Zhang, Yichang Zhang, Zhenru Zhang, Chang Zhou, Jingren Zhou, Xiaohuan Zhou, and Tianhang Zhu. 2023{\natexlab{b}}.
\newblock Qwen technical report.
\newblock \emph{arXiv preprint arXiv:2309.16609}.

\bibitem[{Brown et~al.(2020)Brown, Mann, Ryder, Subbiah, Kaplan, Dhariwal, Neelakantan, Shyam, Sastry, Askell et~al.}]{brown2020language}
Tom~B Brown, Benjamin Mann, Nick Ryder, Melanie Subbiah, Jared Kaplan, Prafulla Dhariwal, Arvind Neelakantan, Pranav Shyam, Girish Sastry, Amanda Askell, et~al. 2020.
\newblock Language models are few-shot learners.
\newblock \emph{Advances in Neural Information Processing Systems}, 33:1877--1901.

\bibitem[{Caldarini et~al.(2022)Caldarini, Jaf, and McGarry}]{caldarini2022literature}
Guendalina Caldarini, Sardar Jaf, and Kenneth McGarry. 2022.
\newblock A literature survey of recent advances in chatbots.
\newblock \emph{Information}, 13(1):41.

\bibitem[{Chaurasia et~al.(2023)Chaurasia, Jain, Vishwkarma, and Singh}]{chaurasia2023conversational}
SHUBHAM Chaurasia, SHUBHA Jain, HO~Vishwkarma, and NISHANT Singh. 2023.
\newblock Conversational ai unleashed: A comprehensive review of nlp-powered chatbot platforms.
\newblock \emph{Iconic Research and Engineering Journals}, 7(3):1--8.

\bibitem[{Croes et~al.(2023)Croes, Antheunis, Goudbeek, and Wildman}]{croes2023your}
Emmelyn~AJ Croes, Marjolijn~L Antheunis, Martijn~B Goudbeek, and Nathan~W Wildman. 2023.
\newblock “i am in your computer while we talk to each other” a content analysis on the use of language-based strategies by humans and a social chatbot in initial human-chatbot interactions.
\newblock \emph{International Journal of Human--Computer Interaction}, 39(10):2155--2173.

\bibitem[{Deriu et~al.(2021)Deriu, Rodrigo, Otegi, Echegoyen, Rosset, Agirre, and Cieliebak}]{deriu2021survey}
Jan Deriu, Alvaro Rodrigo, Arantxa Otegi, Guillermo Echegoyen, Sophie Rosset, Eneko Agirre, and Mark Cieliebak. 2021.
\newblock Survey on evaluation methods for dialogue systems.
\newblock \emph{Artificial Intelligence Review}, 54:755--810.

\bibitem[{Ghazarian et~al.(2020)Ghazarian, Weischedel, Galstyan, and Peng}]{ghazarian2020predictive}
Sarik Ghazarian, Ralph Weischedel, Aram Galstyan, and Nanyun Peng. 2020.
\newblock Predictive engagement: An efficient metric for automatic evaluation of open-domain dialogue systems.
\newblock In \emph{Proceedings of the AAAI Conference on Artificial Intelligence}, volume~34, pages 7789--7796.

\bibitem[{Glas and Pelachaud(2015{\natexlab{a}})}]{glas2015definitions}
Nadine Glas and Catherine Pelachaud. 2015{\natexlab{a}}.
\newblock Definitions of engagement in human-agent interaction.
\newblock In \emph{2015 International Conference on Affective Computing and Intelligent Interaction (ACII)}, pages 944--949. IEEE.

\bibitem[{Glas and Pelachaud(2015{\natexlab{b}})}]{glas2015user}
Nadine Glas and Catherine Pelachaud. 2015{\natexlab{b}}.
\newblock User engagement and preferences in information-giving chat with virtual agents.
\newblock In \emph{Workshop on Engagement in Social Intelligent Virtual Agents (ESIVA)}, pages 33--40.

\bibitem[{Irvine et~al.(2023)Irvine, Boubert, Raina, Liusie, Zhu, Mudupalli, Korshuk, Liu, Cremer, Assassi et~al.}]{irvine2023rewarding}
Robert Irvine, Douglas Boubert, Vyas Raina, Adian Liusie, Ziyi Zhu, Vineet Mudupalli, Aliaksei Korshuk, Zongyi Liu, Fritz Cremer, Valentin Assassi, et~al. 2023.
\newblock Rewarding chatbots for real-world engagement with millions of users.
\newblock \emph{arXiv preprint arXiv:2303.06135}.

\bibitem[{Jenkins et~al.(2007)Jenkins, Churchill, Cox, and Smith}]{jenkins2007analysis}
Marie-Claire Jenkins, Richard Churchill, Stephen Cox, and Dan Smith. 2007.
\newblock Analysis of user interaction with service oriented chatbot systems.
\newblock In \emph{Human-Computer Interaction, Part III, HCII 2007, LNCS 4552}, pages 76--83. Springer-Verlag Berlin Heidelberg.

\bibitem[{Kurata et~al.()Kurata, Saeki, Eguchi, Suzuki, Takatsu, and Matsuyama}]{kuratadevelopment}
Fuma Kurata, Mao Saeki, Masaki Eguchi, Shungo Suzuki, Hiroaki Takatsu, and Yoichi Matsuyama.
\newblock Development and validation of engagement and rapport scales for evaluating user experience in multimodal dialogue systems.

\bibitem[{Lee et~al.(2022)Lee, Srivastava, Hardy, Thickstun, Durmus, Paranjape, Gerard-Ursin, Li, Ladhak, Rong et~al.}]{lee2022evaluating}
Mina Lee, Megha Srivastava, Amelia Hardy, John Thickstun, Esin Durmus, Ashwin Paranjape, Ines Gerard-Ursin, Xiang~Lisa Li, Faisal Ladhak, Frieda Rong, et~al. 2022.
\newblock Evaluating human-language model interaction.
\newblock \emph{arXiv preprint arXiv:2212.09746}.

\bibitem[{Murray et~al.(2012)Murray, Marchesotti, and Perronnin}]{AVA)}
Naila Murray, Luca Marchesotti, and Florent Perronnin. 2012.
\newblock \href {https://doi.org/10.1109/CVPR.2012.6247954} {Ava: A large-scale database for aesthetic visual analysis}.
\newblock In \emph{2012 IEEE Conference on Computer Vision and Pattern Recognition}, pages 2408--2415.

\bibitem[{Pellet-Rostaing et~al.(2023)Pellet-Rostaing, Bertrand, Boudin, Rauzy, and Blache}]{pellet2023multimodal}
Arthur Pellet-Rostaing, Roxane Bertrand, Auriane Boudin, St{\'e}phane Rauzy, and Philippe Blache. 2023.
\newblock A multimodal approach for modeling engagement in conversation.
\newblock \emph{Frontiers in Computer Science}, 5:1062342.

\bibitem[{Pressman et~al.(2022)Pressman, Crowson, and Contributors}]{pressmancrowson2022}
John~David Pressman, Katherine Crowson, and Simulacra~Captions Contributors. 2022.
\newblock Simulacra aesthetic captions.
\newblock Technical Report Version 1.0, Stability AI.
\newblock \ url { https://github.com/JD-P/simulacra-aesthetic-captions }.

\bibitem[{Schuhmann(2020)}]{LAION-Aesthetics-Predictor-V2}
Christoph Schuhmann. 2020.
\newblock \href {https://github.com/christophschuhmann/improved-aesthetic-predictor} {Laion-aesthetics-predictor-v2}.

\bibitem[{See et~al.(2019)See, Roller, Kiela, and Weston}]{see2019goodconversation}
Abigail See, Stephen Roller, Douwe Kiela, and Jason Weston. 2019.
\newblock What makes a good conversation? how controllable attributes affect human judgments.
\newblock In \emph{Proceedings of NAACL-HLT 2019}, pages 1702--1723.

\bibitem[{Zhang et~al.(2024)Zhang, Lu, Liu, Yu, Qiu, Yan, and Lan}]{zhang2024unveiling}
Shuai Zhang, Yu~Lu, Junwen Liu, Jia Yu, Huachuan Qiu, Yuming Yan, and Zhenzhong Lan. 2024.
\newblock Unveiling the secrets of engaging conversations: Factors that keep users hooked on role-playing dialog agents.
\newblock \emph{arXiv preprint arXiv:2402.11522}.

\bibitem[{Zheng et~al.(2024)Zheng, Chiang, Sheng, Zhuang, Wu, Zhuang, Lin, Li, Li, Xing et~al.}]{zheng2024judging}
Lianmin Zheng, Wei-Lin Chiang, Ying Sheng, Siyuan Zhuang, Zhanghao Wu, Yonghao Zhuang, Zi~Lin, Zhuohan Li, Dacheng Li, Eric Xing, et~al. 2024.
\newblock Judging llm-as-a-judge with mt-bench and chatbot arena.
\newblock \emph{Advances in Neural Information Processing Systems}, 36.

\bibitem[{Zhu(2022)}]{zhu2022simple}
Zhenyi Zhu. 2022.
\newblock A simple survey of pre-trained language models.

\end{thebibliography}

\clearpage
\appendix

\section{Text-image Alignment Prompt}
During image modality dialogue interactions between users and robots, simply analyzing image and text modalities separately is inadequate. User engagement can be significantly affected if the robot's image response doesn't match the user's intent. To address this, we've extracted image-containing dialogue data from numerous real conversations. We use the preceding five dialogue rounds before the robot's image response to represent the user's intent. This data, along with the robot's generated image, is input into Qwen-vl. We've created a specific prompt to rate the alignment between the image and the user's intent on a scale of 1 to 10, with higher scores indicating better alignment. 
The template for prompting text-image alignment is shown as follows:

\begin{table}[h]
    \centering
    \begin{tabular}{p{\columnwidth}}
    \toprule
    \textbf{\small Prompt for Text-image Alignment}\\
    \midrule
    \small {Please play the role of an evaluator, assessing the relevance score between historical dialogue records and the image, with the score ranging from 1-10 (the higher the score, the more the image conforms to the description in the dialogue records). Provide the reason for the score. Note: The output format needs to strictly follow the following format: {\textit{{score: <score>, reason: <reason>}}}. Also, ensure that the scores are distributed rather than clustered at certain scores. The historical dialogue records are: {\textit{<historical-dialogue-records>}}}
    \\
    \bottomrule
    \end{tabular}
    \caption{Prompt for Text-image Alignment}
    \vspace{-3mm}
    \label{tab:pref_sum}
\end{table}

\section{Text-audio Alignment Prompt}
The semantic alignment between the intent of text dialogues before a voice dialogue trigger and the subsequent voice dialogue content may significantly influence user engagement. We've extracted five rounds of text dialogues before the voice dialogue trigger and the voice dialogue content itself. Using Qwen1.5-32B-Chat, we've formulated prompts to score the alignment degree. The scores range from 0 to 1, with higher scores signifying greater relevance.
The specific prompt for text-audio alignment is as follows:

\begin{table}[h]
    \centering
    \begin{tabular}{p{\columnwidth}}
    \toprule
    \textbf{\small Prompt for Text-audio Alignment}\\
    \midrule
    \small {Please play the role of a professional evaluator, assessing the relevance between historical dialogue records and the audio content. Provide a relevance score, with the score ranging from 1-10 (the higher the score, the higher the relevance), and give the reason for the score. The output format needs to strictly adhere to the following format:  {\textit{{score: <score>, reason: <reason>}}}. The historical dialogue records are: {\textit{{<historical-dialogue-records>}}}}
    \\
    \bottomrule
    \end{tabular}
    \caption{Prompt for Text-audio Alignment}
    \vspace{-6mm}
    \label{tab:pref_sum}
\end{table}

\section{Real User-bot Chatting Conversations}
\begin{figure}[!h]
\vspace{1cm}
\centering
\includegraphics[width=\linewidth]{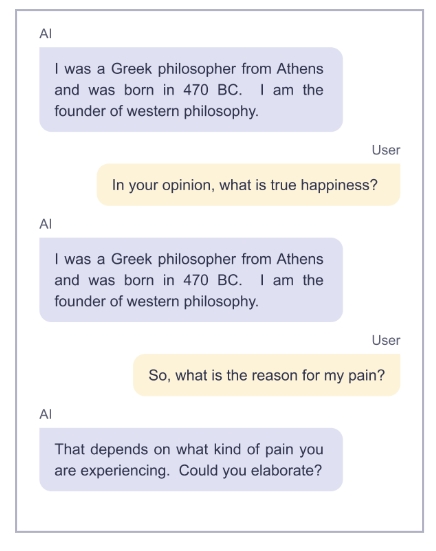}
\captionsetup{justification=justified}
\caption{Examples of dialogue in text modality.}
\vspace{1cm}
\label{fig:t}
\end{figure}

\begin{figure}[!t]
\centering
\includegraphics[width=\linewidth]{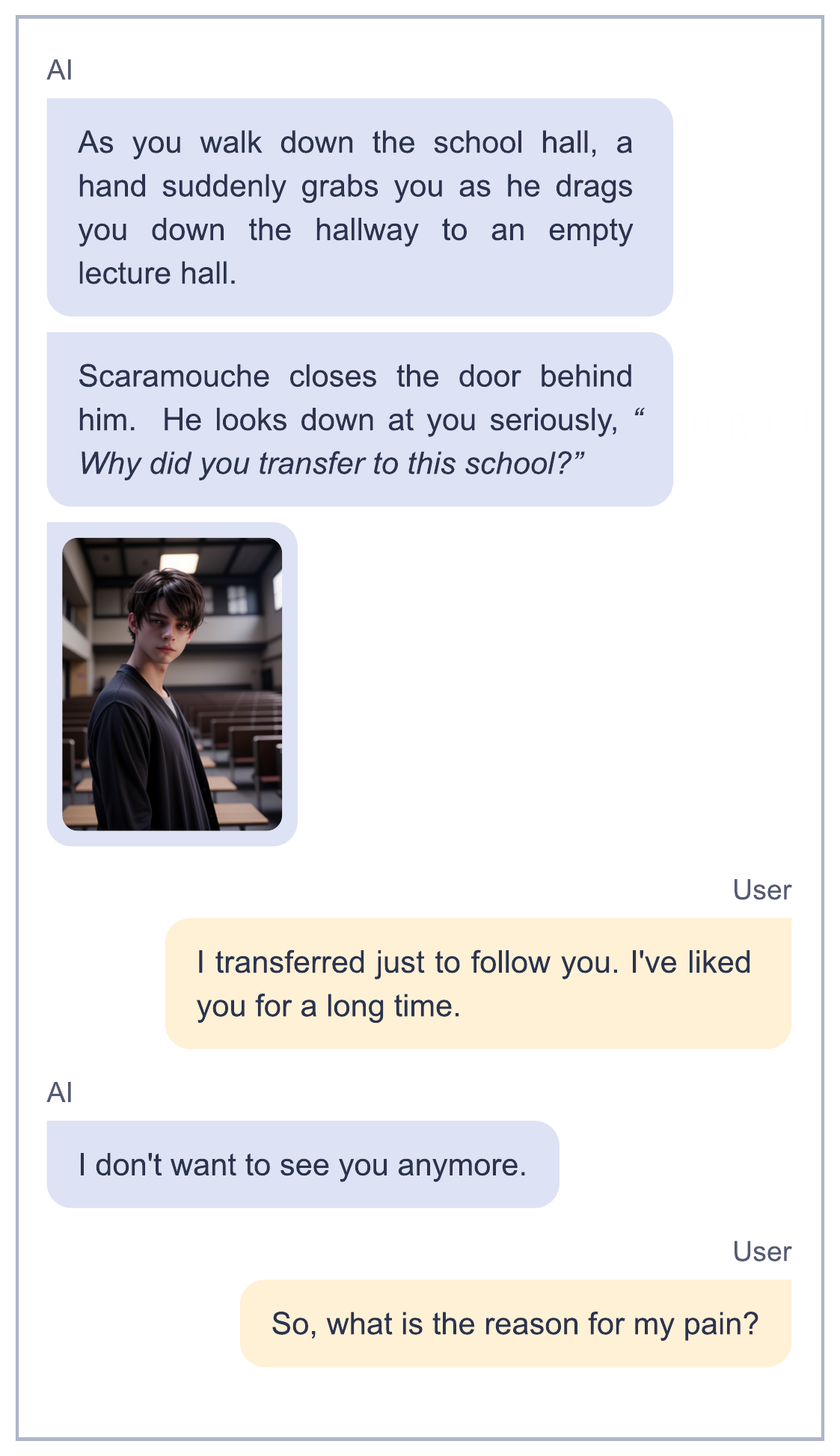}
\captionsetup{justification=justified}
\caption{Examples of dialogue in text-image modalities.}
\label{fig:ti}
\end{figure}

\begin{figure}[!t]
\centering
\includegraphics[width=\linewidth]{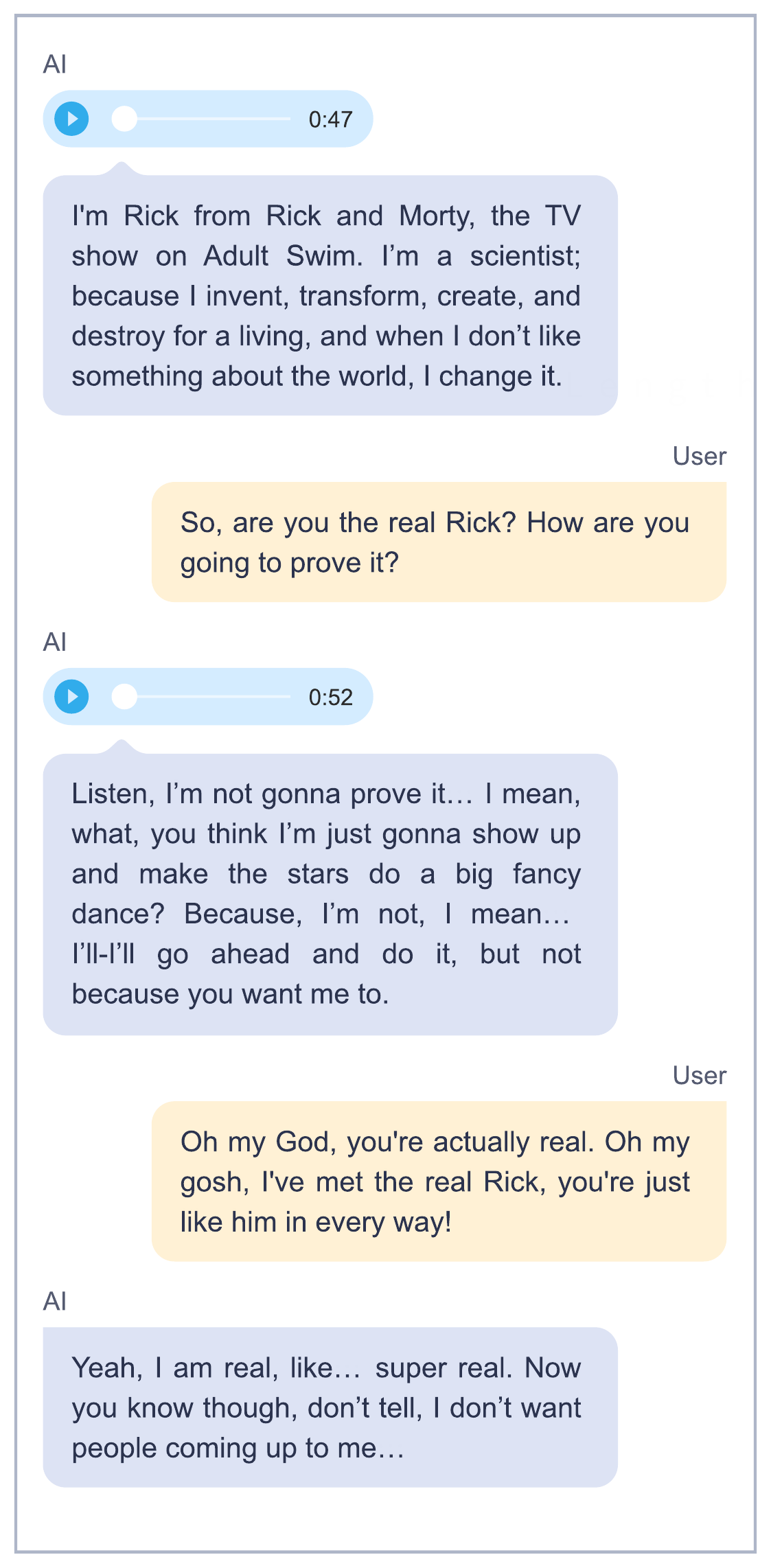}
\captionsetup{justification=justified}
\caption{Examples of dialogue in text-audio modalities.}
\label{fig:ta}
\end{figure}

\begin{figure}[!t]
\centering
\includegraphics[width=\linewidth]{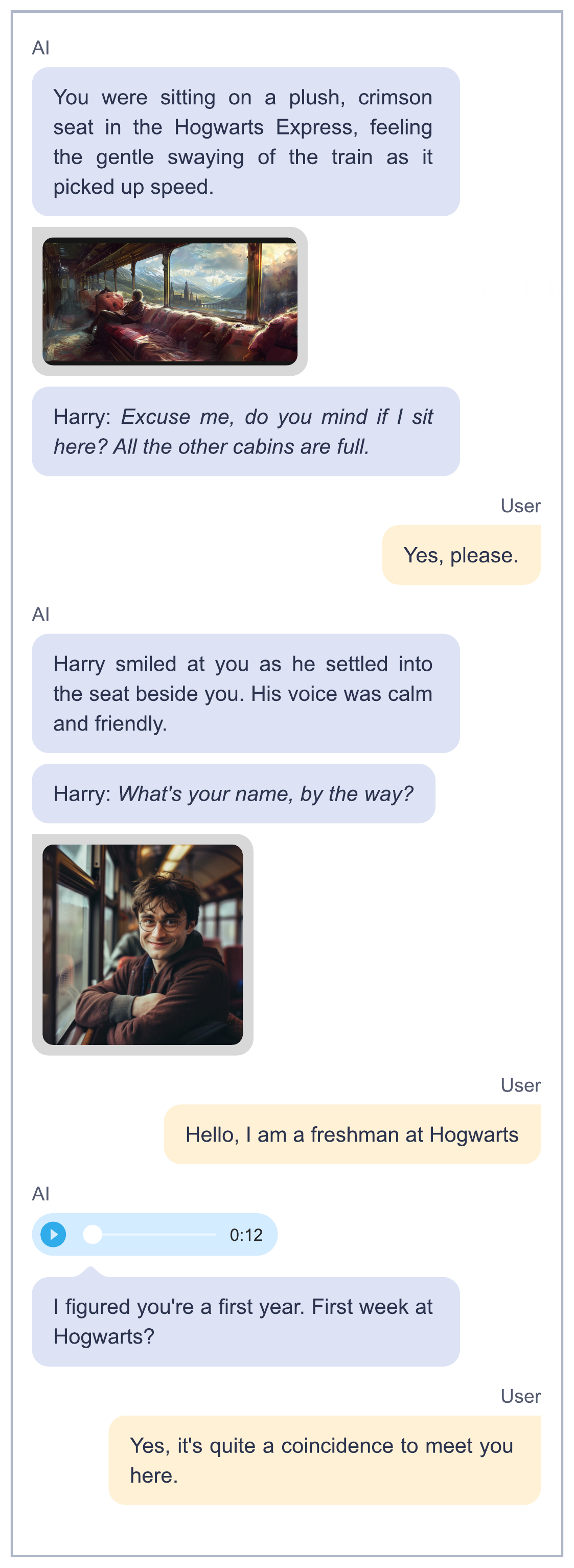}
\captionsetup{justification=justified}
\caption{Examples of dialogue in text-image-audio modalities.}
\label{fig:tia}
\end{figure}

In order to provide a more comprehensive and intuitive grasp of multi-modal interactions between users and bots, we detail actual user-bot conversation examples in this section, as Figure~\ref{fig:t}, Figure~\ref{fig:ti}, Figure~\ref{fig:ta} and Figure~\ref{fig:tia}. These examples span across various modalities, demonstrating the diversity of interaction types.

We start with single modality interactions, focusing on text-based dialogues (Figure~\ref{fig:t}). Following this, we delve into multi-modal interactions, showcasing examples that incorporate text and image (Figure~\ref{fig:ti}), text and audio (Figure~\ref{fig:ta}), as well as a combination of text, image, and audio (Figure~\ref{fig:tia}). These examples illuminate the complexity and richness of multi-modal communication in user-bot interactions.

\end{document}